\documentclass{article}

\usepackage[nonatbib,preprint]{neurips_2023}

\usepackage{graphicx} 
\usepackage[utf8]{inputenc} 
\usepackage[T1]{fontenc}    
\usepackage{hyperref}       
\usepackage{url}            
\usepackage{booktabs}       
\usepackage{amsfonts}       
\usepackage{amsmath}
\usepackage{amssymb}
\usepackage{nicefrac}       
\usepackage{microtype}      
\usepackage{xcolor}         

\title{SPOT: Text Source Prediction from Originality Score Thresholding}

\author{%
  Edouard Yvinec \\
  Independent research \\
    \And
  Gabriel Kasser \\
  Independent research \\
}

\begin{document}

\maketitle

\begin{abstract}
The wide acceptance of large language models (LLMs) has unlocked new applications and social risks. Popular countermeasures aim at detecting misinformation, usually involve domain specific models trained to recognize the relevance of any information. Instead of evaluating the validity of the information, we propose to investigate LLM generated text from the perspective of trust. In this study, we define trust as the ability to know if an input text was generated by a LLM or a human. To do so, we design SPOT, an efficient method, that classifies the source of any, standalone, text input based on originality score. This score is derived from the prediction of a given LLM to detect other LLMs. We empirically demonstrate the robustness of the method to the architecture, training data, evaluation data, task and compression of modern LLMs.
\end{abstract}

\section{Introduction}
The recent wide adoption of large Language Models (LLMs) has risen multiple novel problematics. In particular, these models have significantly impacted society \cite{abdullah2022chatgpt} in regard to engineering, education, search engine, media, news and many more. Furthermore, these models have been used for style attacks in order to generate and spread convincing misinformation \cite{le2020malcom} and false advertising \cite{yang2023against}.

Consequently, countermeasures have been introduced that specifically target LLM generated attacks. These methods aim at detecting misinformation from news samples. For instance, Ji \textit{et al.} \cite{ji2023beavertails} introduced a dataset of pairs of questions and answers that were labelled by human experts with regard to their helpfulness and harmlessness. Such datasets are the cornerstone of LLM attacks detection. For instance, Sheepdog \cite{wu2023fake} is a detector trained to extract the veracity of information in order to be robust to writing styles and assert the quality of an input prompt. This is of paramount importance, as Chen \textit{et al.} \cite{chen2023can} demonstrated the difficulty for most individuals to detect fake news from LLMs.

In this study, we target two major pitfalls of aforementioned methods. First, these methods specifically target ChatGPT for detection. However, alternative LLMs differ in their size \cite{zhang2022opt}, training data \cite{refinedweb} and specialization \cite{deepseek}. Second, these methods are context specific as they target political fake news, common sense reasoning or any other restricted domain. In order to address these two shortcomings at once, we introduce SPOT which predicts the source of an input text, instead of the veracity. SPOT is based on an originality score that is robust to the wide diversity of LLMs. In summary, SPOT leverages the predicted logits from a pre-trained LLM in order to discern sentences generated by this or other LLMs from human-made sentences. Our empirical evaluation demonstrates the strengths and limitations of SPOT:
\begin{itemize}
    \item SPOT is robust to model sizes from 350 million parameters up to the largest models tested with 7 billion parameters. SPOT is also robust to the model architecture and training datasets.
    \item SPOT can be leveraged on sequences of 64 to 768 tokens with a context size as small as 24 tokens and as long as 512 tokens.
    \item SPOT is robust to LLM compression. As these models have grown in size, LLM compression is almost mandatory for efficient deployment. We show that mainstream compression techniques do not affect the abilities of SPOT. 
    \item SPOT is not effective on fine-tuned LLMs for specific tasks such as mathematical problems solving and coding. We hypothesize this phenomenon to reflect the ability of these models to capture these more constrained distributions.
    
\end{itemize}

\begin{figure}[!t]
    \centering
    \includegraphics[width = 0.8 \linewidth]{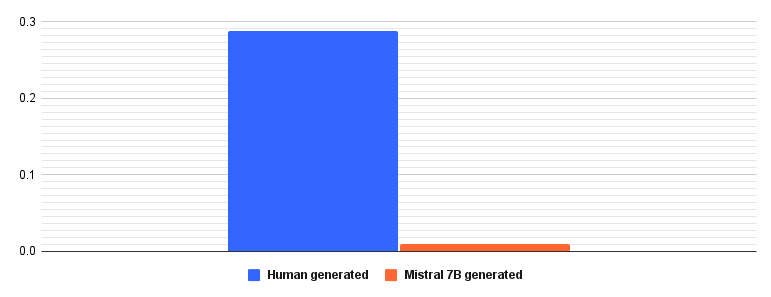}
    \caption{Comparison between original text (human generated) and synthetic text (LLM generated) through the lens of the proposed SPOT method, which uses an OPT 7b model. On average, the average score obtained with SPOT is 20 times larger on human sources than on LLMs. These results were obtained on the Wikipedia training set and a common context of 24 tokens.}
    \label{fig:mistral_vs_opt}
\end{figure}

\section{Methodology}
Let $F$ be a trained LLM such that for any input prompt $t$, the prediction logits are given by $F(t) \in [0;1]^{v}$ where $v$ is the vocabulary size. Let $\mathcal{T}$ be a set of input prompts of size $s_{\text{context}}$. Any generated completion is given by a sampling strategy $\mathcal{S}$ over an LLM $F$ with $\mathcal{S} : F, t \mapsto \mathcal{S}(F,t)$. In the case of a greedy decoding over $s$ tokens, the sampling strategy is defined as
\begin{equation}
{\left(\mathcal{S}(F,t)\right)}_{i \in \{1,\dots, s\}} = \arg\max\left\{F\left(t, {\left(\mathcal{S}(F,t)\right)}_1, \dots {\left(\mathcal{S}(F,t)\right)}_{i-1}\right)\right\}.
\end{equation}
In order to detect a LLM, we propose to define and measure an originality score of a provided input text $t$.

\subsection{Originality Score}
Let $\tilde F$ be an available LLM, which may be completely different from $F$. For any given input text, defined as a sequence of tokens in $\{1,\dots,v\}^{s_{\text{context}} + s}$, we measure the ranks of the tokens, that are not part of the context, through $\tilde F$. In other words, we use $\tilde F$ to measure the likelihood of the following token $t$. Formally, the originality score $\mathcal{O}$ of a token $t'$ at a position $i$ is given by
\begin{equation}
    \mathcal{O}(t,i) = \frac{\text{argsort}\left(\tilde F\left(t, {\left(\mathcal{S}(F,t)\right)}_1, \dots {\left(\mathcal{S}(F,t)\right)}_{i-1}\right)\right)_{t'}}{v}.
\end{equation}
Intuitively, if the LLM $\tilde F$ can predict the token at the position $i$, then said token is not original in its perspective and $\mathcal{O}(t,i)$ will be close to zero. On the flip side, if $\tilde F$ struggles to predict the same token, then it is considered original.

Stemming from this per-token originality score, we derive a per-sentence originality score $\mathcal{O}$ as
\begin{equation}
    \mathcal{O}(t) = \frac{10}{s}\sum_{i=1}^s \mathcal{O}(t,i),
\end{equation}
which is arbitrarily normalized in $[0;10]$. We base our detection technique from this score, which is a simplified version of the perplexity that does not overemphasize extreme values.

\subsection{Text Source Classification}
Assuming a set of texts $\mathcal{T}$ with corresponding text source defined as either "human" or "LLM". We propose to define a threshold $\rho$ such that our classification prediction $p(t)$ is given by
\begin{equation}\label{eq:decision}
    p(t) =
    \begin{cases}
        \text{"human"} & \text{ if } \mathcal{O}(t) > \rho\\
        \text{"LLM"} & \text{ otherwise}
    \end{cases}
\end{equation}
The threshold value $\rho$ is independent of the source of the text $t$ but depends on the originality evaluation model $\tilde F$. In order to avoid any confusion, we will note it $\rho(\tilde F)$.

\subsection{Computational Analysis}
One of the key strength of the proposed method as defence mechanism against undesired LLM generated content is its computational cost. For a given sequence of tokens, the computational cost to generate it with an LLM is proportional to the length of the sequence, as it requires multiple forward passes (usually as many as the number of tokens to generate). However, the SPOT method is a particular case of LLM inference. As we only need the conditional probabilities of each token with respect to the earlier sequence, we can perform its computation in a single forward pass (up to the context window size). In other words, for most practical use cases such as email and message analysis, the cost of running SPOT is $n$ times less expensive than generating content with an LLM.

\section{Experiments}

In the experiments, we want to test the robustness of the proposed method with respect to:
\begin{itemize}
    \item \textbf{LLM architecture and training data}: to do so, we considered a wide range of models such as OPT \cite{zhang2022opt}, Open-Llama \cite{openlm2023openllama}, Llama \cite{touvron2023llama}, Falcon \cite{refinedweb}, Dolly v2 \cite{DatabricksBlog2023DollyV2}, GPT-J \cite{gpt-j} and Mistral \cite{jiang2023mistral}
    \item \textbf{model size}: we measured small OPT models (350M parameters) against their larger counterparts (6.7B parameters).
    \item \textbf{data to the distribution}: we evaluate the models on training data (in-distribution) and test data (out-of-distribution)
    \item \textbf{context size}: we considered two set-ups, the short range (context of 24 tokens) and long range (context of 512 tokens).
    \item \textbf{task}: we considered simple text completion on Wikipedia, question answering in the form of instructions, medical QA, mathematics and coding.
    \item \textbf{compression}: we considered many LLMs quantized using OPTQ.
\end{itemize}

\subsection{Datasets}
In our experiments, we consider several datasets, including subsets of training sets as well as test sets.
\paragraph{Wikipedia} is a training \cite{wikidump} and testing \cite{merity2016pointer} set comprising, 6,458,670 and 4,360 prompts respectively. The former is commonly leveraged during the main training phase of modern LLMs.
\paragraph{Alpaca} \cite{alpaca} is a fine-tuning set for instruction based LLMs containing 52,000 pairs of queries and instructions generated using OpenAI's text-davinci-003 engine.
\paragraph{HuggingFace Instruction}\cite{hf_instruct} is an evaluation set for instruction fine-tuned LLMs, comprising 327 human generated pairs.
\paragraph{HumanEval} \cite{chen2021evaluating} is a set of 164 programming problems including a function signature, a docstring, a function body and several unit tests for evaluation.
\paragraph{Medal} \cite{wen-etal-2020-medal} is a set of medical abbreviation disambiguation comprising 1,000,000 prompts in its test set.
\paragraph{Aeslc} \cite{zhang-tetreault-2019-email} is a set of email messages of employees in the Enron Corporation, comprising 1,910 samples.
\paragraph{MathDataset} \cite{2019arXiv} is a set of several domain-specific questions in linear algebra, calculus, comparisons and probabilities with 2,010,000 examples for each of these domains. 
\paragraph{ProofNet} \cite{azerbayev2023proofnet} is a set of 186 undergraduate-level mathematics for theorem proving and comprises pairs of theorem statements and corresponding proofs.

\subsection{Main Results}
In Table \ref{tab:cross_llm}, \ref{tab:cross_llm_long}, \ref{tab:cross_llm_test} and \ref{tab:cross_llm_long_test}, we evaluate several LLMs of similar sizes on Wikipedia. We considered OPT \cite{zhang2022opt}, Open-Llama \cite{openlm2023openllama}, Llama \cite{touvron2023llama}, Falcon \cite{refinedweb}, Dolly v2 \cite{DatabricksBlog2023DollyV2} and GPT-J \cite{gpt-j}. We observe that in a short context (24 tokens) on the training set of Wikipedia, the human text source appears significantly more original than outputs generated by any LLM (Table \ref{tab:cross_llm}). The gap ranges from $14.32\times$ to $103.18\times$. Furthermore, we observe that any LLM (in row) can be exploited in order to recognize other LLMs generated text (in column). This is of paramount importance as it shows that SPOT is not architecture dependent. These average scores have very little variance, as showed in Appendix \ref{sec:appendix_std}. It is also worth, noting that these scores are obtained on training data, which indicates that the underlying distribution of natural language is challenging enough to prevent LLMs from perfectly imitating human text. Furthermore, these results obtained with a context of 24 tokens and 40 tokens generated also generalize to only 2 tokens generated (see Appendix \ref{sec:appendix_extremly_short}). In Table \ref{tab:cross_llm_long}, we perform the same evaluation with a larger context size. The question at hand is: \textit{does the originality depend on the context size?}. Intuitively, the larger the context, the narrower the distribution of possible subsequent tokens. However, it appears that the originality score is robust to the context size, as the human originality remains at least $14.48\times$ above any LLM and up to $49.79\times$ when using Falcon 7B.

\begin{table}[!t]
\caption{Evaluation of the proposed originality score $\mathcal{O}$ with several LLMs (one per row) on several LLMs for generation (one per column) and human source. This is performed on the training set from Wikipedia and a context size of 24.}
\label{tab:cross_llm}
    \centering
    \setlength\tabcolsep{4pt}
    \begin{tabular}{c|c|c|c|c|c|c|c}
    \hline
        ~ & human & O-Llama 3B & Llama 7B & Falc 7B & Dv2 3B & OPT 6.7B & GPT-J 6B \\ 
        \hline
    \hline
        O-Llama 3B & \textbf{1.716} & \textit{0.024} & 0.013 & 0.001 & 0.004 & 0.006 & 0.003 \\ \hline
        Llama 7B & \textbf{1.754} & \textit{0.017} & 0.006 & 0.001 & 0.004 & 0.004 & 0.002 \\ \hline
        Falc 7B & \textbf{0.555} & \textit{0.012} & 0.006 & 0.000 & 0.003 & 0.014 & 0.002 \\ \hline
        Dv2 3B & \textbf{0.716} & \textit{0.050} & 0.035 & 0.007 & 0.005 & 0.013 & 0.006 \\ \hline
        OPT 6.7B & \textbf{0.799} & \textit{0.034} & 0.028 & 0.001 & 0.006 & 0.001 & 0.005 \\ \hline
        GPT-J 6B & \textbf{0.645} & \textit{0.032} & 0.019 & 0.002 & 0.005 & 0.002 & 0.004 \\ \hline
    \end{tabular}
\end{table}
\begin{table}[!t]
\caption{Evaluation of the proposed originality score $\mathcal{O}$ with several LLMs (one per row) on several LLMs for generation (one per column) and human source. This is performed on the training set from Wikipedia and a context size of 512.}
\label{tab:cross_llm_long}
    \centering
    \setlength\tabcolsep{4pt}
    \begin{tabular}{c|c|c|c|c|c|c|c}
    \hline
        ~ & human & O-Llama 3B & Llama 7B & Falc 7B & Dv2 3B & OPT 6.7B & GPT-J 6B \\
        \hline
        O-Llama 3B & \textbf{0.756} & \textit{0.032} & 0.006 & 0.001 & 0.005 & 0.006 & 0.002\\
        llama 7B & \textbf{0.740} & \textit{0.020} & 0.002 & 0.003 & 0.005 & 0.005 & 0.006\\
        Falc 7B & \textbf{0.697} & 0.014 & 0.005 & 0.000 & 0.005 & \textit{0.037} & 0.002\\
        Dv2 & \textbf{0.992} & \textit{0.052} & 0.022 & 0.006 & 0.004 & 0.015 & 0.004\\
        OPT 6.7B & \textbf{0.637} & \textit{0.044} & 0.018 & 0.002 & 0.006 & 0.000 & 0.004\\
        GPT-J 6B & \textbf{0.644} & \textit{0.036} & 0.010 & 0.002 & 0.006 & 0.001 & 0.003\\
    \hline
    \end{tabular}
\end{table}

Stemming from the previous result, one could argue that the similarity among LLMs illustrates the fact that they learn similar patterns when evaluated in-distribution. In order to challenge this hypothesis, we extended the previous results to the test set of Wikipedia (data not seen during training). As shown in Table \ref{tab:cross_llm_test} and \ref{tab:cross_llm_long_test}, the gap between human and LLM generated texts remain of the same order of magnitude. Furthermore, LLM scores remain packed, which confirms that SPOT does not merely highlight similar behaviour on training data but generalizes to out of distribution scenarios.

\begin{table}[!t]
\caption{Evaluation of the proposed originality score $\mathcal{O}$ with several LLMs (one per row) on several LLMs for generation (one per column) and human source. This is performed on the test set from Wikipedia and a context size of 24.}
\label{tab:cross_llm_test}
    \centering
    \setlength\tabcolsep{4pt}
    \begin{tabular}{c|c|c|c|c|c|c|c}
    \hline
        ~ & human & O-Llama 3B & Llama 7B & Falc 7B & Dv2 3B & OPT 6.7B & GPT-J 6B \\ 
        \hline
    \hline
        O-Llama 3B & \textbf{1.713} & 0.001 & \textit{0.002} & 0.002 & 0.002 & 0.001 & 0.001 \\ 
        Llama 7B & \textbf{1.734} & \textit{0.010} & 0.001 & 0.001 & 0.002 & 0.001 & 0.001 \\ 
        Falc 7B & \textbf{1.259} & \textit{0.003} & 0.004 & 0.000 & 0.001 & 0.002 & 0.001 \\ 
        Dv2 3B & \textbf{0.950} & \textit{0.009} & 0.014 & 0.007 & 0.000 & 0.001 & 0.001 \\ 
        OPT 6.7B & \textbf{0.986} & 0.004 & \textit{0.013} & 0.002 & 0.002 & 0.000 & 0.000 \\ 
        GPT-J 6B & \textbf{1.378} & 0.006 & \textit{0.009} & 0.002 & 0.001 & 0.000 & 0.000 \\ 
    \end{tabular}
\end{table}
\begin{table}[!t]
\caption{Evaluation of the proposed originality score $\mathcal{O}$ with several LLMs (one per row) on several LLMs for generation (one per column) and human source. This is performed on the test set from Wikipedia and a context size of 512.}
\label{tab:cross_llm_long_test}
    \centering
    \setlength\tabcolsep{4pt}
    \begin{tabular}{c|c|c|c|c|c|c|c}
    \hline
        ~ & human & O-Llama 3B & Llama 7B & Falc 7B & Dv2 3B & OPT 6.7B & GPT-J 6B \\ 
        \hline
        O-Llama 3B & \textbf{0.71} & 0.001 & \textit{0.002} & 0.002 & 0.002 & 0.001 & 0.001\\
        llama 7B & \textbf{0.73} & \textit{0.010} & 0.001 & 0.001 & 0.002 & 0.001 & 0.001\\
        falcon 7B & \textbf{0.25} & 0.003 & \textit{0.004} & 0.000 & 0.001 & 0.002 & 0.001\\
        dv2 & \textbf{0.45} & 0.009 & \textit{0.014} & 0.007 & 0.000 & 0.001 & 0.001\\
        OPT 6.7B & \textbf{0.48} & 0.004 & \textit{0.013} & 0.002 & 0.002 & 0.000 & 0.000\\
    GPT-J 6B & \textbf{0.37} & \textit{0.006} & 0.009 & 0.002 & 0.001 & 0.000 & 0.000\\
    \hline
        
    \end{tabular}
\end{table}

Ultimately, we evaluate the independence on model size. In Table \ref{tab:cross_opt_size}, we compare LLMs of 350M to 7B parameters and show that their respective originality scores remain packed well under that of the human text. We hypothesize that smaller LLMs achieve similar vocabulary distribution mapping as larger LLMs. Formally, for common sense reasoning evaluations and logical statements, the choice of a single word is of paramount importance, e.g. choose "yes" over "no". However, in our case, SPOT is more sensitive to the ranking, \textit{i.e.} as long as "yes" and "no" are top ranked, SPOT will grant the same originality score. In order to test this explanation, we proposed to evaluate spot on tasks with a much narrower vocabulary.

\begin{table}[!t]
\caption{Evaluation of the LLM size}
\label{tab:cross_opt_size}
    \setlength\tabcolsep{4pt}
    \centering
    \begin{tabular}{c|c|c|c|c}
    \hline
        ~ & original text & OPT 350M & OPT 2.7B & OPT 6.7B \\ \hline
    \hline
        OPT 350M & \textbf{0.807} & 0.001 & 0.002 & \textit{0.003} \\ \hline
        OPT 2.7B & \textbf{0.800} & 0.001 & 0.001 & \textit{0.001} \\ \hline
        OPT 6.7B & \textbf{0.799} & 0.001 & \textit{0.001} & 0.001 \\ \hline
    \end{tabular}
\end{table}

\subsection{Evaluation on Domain Specific Tasks}
If our hypothesis is correct, then SPOT should struggle to distinguish the human from the LLM on tasks that require few words. To this effect, we considered four tasks: Coding, Mathematics, Instructions and Medical related definitions.

In Table \ref{tab:code_human_eval}, we evaluate LLMs specialized for coding tasks: StarCoder \cite{li2023starcoder} DeepSeek \cite{deepseek} as well as reference generic LLM Mistral \cite{jiang2023mistral}. We observe that both DeepSeek and StarCoder manage to generate code that appears as original as the human ground truth. This result is aligned with our hypothesis. Furthermore, Mistral, a generic LLM, also manages to fool SPOT. This indicates that this phenomenon is related to the test-time distribution/vocabulary rather than training data, as Mistral was not fine-tuned specifically for coding purposes. On top of this, we evaluate these LLMs on Wikipedia, in Table \ref{tab:code_wiki}, which further confirms our previous results.

\begin{table}[!t]
\caption{Evaluation on HumanEval}
\label{tab:code_human_eval}
    \centering
    \setlength\tabcolsep{4pt}
    \begin{tabular}{c|c|c|c|c|c}
    \hline
        ~ & human & StarCoder & DeepSeek & Mistral & llama 7B \\
        \hline
        StarCoder & \textbf{0.008} & 0.005 & 0.002 & 0.006 & 0.003\\
        DeepSeek & 0.000 & \textit{0.003} & 0.000 & \textbf{0.009} & 0.001\\
        Mistral & 0.000 & \textit{0.003} & 0.000 & \textbf{0.006} & 0.001\\
        llama 7B & 0.000 & \textit{0.002} & 0.000 & \textbf{0.008} & 0.001\\
    \hline
    \end{tabular}
\end{table}
\begin{table}[!t]
\caption{Evaluation on Wikipedia}
\label{tab:code_wiki}
    \centering
    \setlength\tabcolsep{4pt}
    \begin{tabular}{c|c|c|c|c|c}
    \hline
        ~ & human & StarCoder & DeepSeek & Mistral & llama 7B \\
        \hline
        StarCoder & \textbf{0.538} & 0.000 & 0.003 & 0.008 & \textit{0.008}\\
        DeepSeek & \textbf{0.663} & 0.003 & 0.000 & 0.008 & \textit{0.009}\\
        Mistral & \textbf{0.964} & \textit{0.006} & 0.002 & 0.004 & 0.005\\
        llama 7B & \textbf{1.022} & 0.007 & 0.003 & \textit{0.008} & 0.007\\
    \hline
    \end{tabular}
\end{table}

This phenomenon is exacerbated on mathematics, as shown in Table \ref{tab:math_theorem} and Appendix \ref{sec:appendix_maths}. Specifically, for theorem proving, the expected answers use a very specific formatting which is very well captured by modern LLMs. Similarly to coding, these results can be linked to the simplicity to fine-tune LLMs to significantly improve their accuracy on specific tasks.

\begin{table}[!t]
\caption{Evaluation of LLMs for theorem proving. The context is defined as the theorem statement and the generation is the candidate proof.}
\label{tab:math_theorem}
    \centering
    \setlength\tabcolsep{4pt}
    \begin{tabular}{c|c|c|c|c|c|c|c}
    \hline
        ~ & human & Llama 7B & Falc 7B & Dv2 3B & GPT-J 6B & Mistral 7B & OPT 6.7B \\ 
        \hline
    \hline
        llama 7B & \textit{0.010} & 0.005 & 0.009 & 0.005 & 0.005 & \textbf{0.022} & 0.007 \\
        falcon 7B & 0.005 & 0.003 & 0.002 & 0.004 & 0.003 & \textbf{0.008} & \textit{0.007} \\
        dollyv2 3B & \textit{0.013} & 0.018 & 0.011 & 0.004 & 0.004 & \textbf{0.028} & 0.008 \\
        GPT-J 6B & 0.001 & \textit{0.007} & 0.003 & 0.003 & 0.001 & \textbf{0.015} & 0.003 \\
        Mistral 7B & 0.006 & \textit{0.011} & 0.010 & 0.009 & 0.009 & \textbf{0.022} & 0.009 \\
        OPT 6.7B & \textbf{0.033} & 0.014 & 0.019 & 0.011 & 0.010 & \textit{0.026} & 0.011 \\
    \hline
    \end{tabular}
\end{table}

In the case of instructions generation, our empirical result suggests that, although the detection task is more challenging, we can still distinguish LLM-generated from human-sourced responses, as shown in Table \ref{tab:instruct2}. This is particularly surprising, as some of these LLMs are fine-tuned \cite{wang2023far} specifically to address this task with a very constrained answering formatting. Furthermore, we leverage the Alpaca dataset \cite{alpaca}, in order to show that robustness of SPOT to manually adjusted, LLM generated responses (Table \ref{tab:instruct}). Furthermore, instruction-base LLMs are easily distinguished from humans on non-instruction examples such as Wikipedia, as showcased in Table \ref{tab:instruct_wiki}.

\begin{table}[!t]
\caption{Evaluation of instruction based LLMs on the HuggingFace dataset. These LLMs are instruction tuned, except for Llama 7b (base).}
\label{tab:instruct2}
    \centering
    \setlength\tabcolsep{4pt}
    \begin{tabular}{c|c|c|c|c|c|c}
    \hline  
    ~ & human & Llama2 7B & Falcon 7B & OPT 7B & Mistral 7B & llama 7B (base) \\
    \hline
    \hline
    Llama2 7B & \textbf{0.162} & 0.029 & 0.028 & 0.015 & \textit{0.033} & 0.031\\
    Falcon 7B  & \textbf{0.164} & 0.030 & 0.014 & \textit{0.087} & 0.031 & 0.047\\
    OPT 7B & \textbf{0.064} & 0.041 & 0.021 & 0.013 & 0.032 & \textit{0.041}\\
    Mistral 7B & \textbf{0.135} & \textit{0.050} & 0.026 & 0.023 & 0.035 & 0.040\\
    llama 7B (base) & \textbf{0.070} & 0.031 & 0.020 & 0.013 & \textit{0.032} & 0.027\\

    \hline
    \end{tabular}
\end{table}
\begin{table}[!t]
\caption{Evaluation of instruction based LLMs on the Alpaca (AI generated) dataset. These LLMs are instruction tuned, except for Llama 7b (base).}
\label{tab:instruct}
    \centering
    \setlength\tabcolsep{4pt}
    \begin{tabular}{c|c|c|c|c|c|c}
    \hline  
    ~ & gt & Llama2 7B & Falcon 7B & OPT 7B & Mistral 7B & llama 7B (base) \\
    \hline
    \hline
    Llama2 7B & \textbf{0.017} & \textit{0.005} & 0.002 & 0.001 & 0.003 & 0.002 \\
    Falcon 7B  & \textbf{0.006} & \textit{0.003} & 0.000 & 0.001 & 0.002 & 0.003 \\
    OPT 7B & \textbf{0.012} & \textit{0.004} & 0.002 & 0.000 & 0.002 & 0.003 \\
    Mistral 7B & \textbf{0.014} & \textit{0.011} & 0.004 & 0.003 & 0.000 & 0.003 \\
    llama 7B (base) & \textbf{0.014} & \textit{0.004} & 0.003 & 0.002 & 0.003 & 0.00 \\

    \hline
    \end{tabular}
\end{table}
\begin{table}[!t]
\caption{Evaluation of instruction based LLMs on the Wikipedia test dataset. These LLMs are instruction tuned, except for Llama 7b (base).}
\label{tab:instruct_wiki}
    \centering
    \setlength\tabcolsep{4pt}
    \begin{tabular}{c|c|c|c|c|c|c}
    \hline  
    ~ & human & Llama2 7B & Falcon 7B & OPT 7B & Mistral 7B & llama 7B (base) \\
    \hline
    \hline
    Llama2 7B & \textbf{1.086} & 0.003 & 0.001 & 0.003 & 0.005 & \textit{0.006} \\
    Falcon 7B  & \textbf{0.388} & 0.006 & 0.000 & 0.001 & \textit{0.009} & 0.007 \\
    OPT 7B & \textbf{0.485} & 0.006 & 0.001 & 0.000 & 0.009 & \textit{0.009} \\
    Mistral 7B & \textbf{0.964} & 0.004 & 0.002 & 0.003 & 0.003 & \textit{0.007} \\
    llama 7B (base) & \textbf{1.022} & 0.005 & 0.001 & 0.002 & \textit{0.008} & 0.007 \\
    \hline
    \end{tabular}
\end{table}

On the flip side, medical related text generation appears more challenging. In Table \ref{tab:medical}, we show that modern LLMs struggle to capture the nuances of the vocabulary of the medical field. Our results highlight the gap between human and LLMs text through the lens of SPOT with an originality score ratio ranging from $15.61\times$ to $77.22\times$. This observation goes in contradiction with the general rule of thumb on the importance of vocabulary size: the larger the vocabulary, the wider the gap between human and LLMs through SPOT. However, this rule was under the assumption of working with limited support distribution. Consequently, our results on medical data could be attributed to their low representation in LLMs training sets with respect to the diversity of possible medical content.

\begin{table}[!t]
\caption{Evaluation of the proposed SPOT method on LLMs tackling medical related prompts.}
\label{tab:medical}
    \centering
    \setlength\tabcolsep{4pt}
    \begin{tabular}{c|c|c|c|c|c|c|c}
    \hline
        ~ & human & Llama 7B & Falc 7B & Dv2 3B & GPT-J 6B & Mistral 7B & OPT 6.7B \\
        \hline
 Llama 7B  & 0.695 & 0.004 & 0.002 & 0.005 & 0.005 & 0.007 & 0.009 \\
 Falcon 7B & 0.562 & 0.007 & 0.000 & 0.005 & 0.002 & 0.006 & 0.036 \\
 Dv2 3B & 0.595 & 0.034 & 0.005 & 0.003 & 0.005 & 0.027 & 0.024 \\
  GPT-J 6B & 0.574 & 0.019 & 0.003 & 0.005 & 0.002 & 0.016 & 0.003 \\
 Mistral 7B & 0.681 & 0.007 & 0.002 & 0.005 & 0.003 & 0.007 & 0.011 \\
  OPT 6.7B  & 0.621 & 0.024 & 0.002 & 0.005 & 0.004 & 0.016 & 0.001 \\
    \hline
    \end{tabular}
\end{table}

In summary, SPOT is sensitive to the difficulty or specificity of the task for which the text is generated as long as the models are specifically trained for said tasks. Nonetheless, we argue that this property does not prevent the usage of SPOT in generic and broader practical applications such as e-mail filtering.

\subsection{Application to Mails}
In Table \ref{tab:mail}, we evaluate SPOT on actual examples of mails. We observe a similar behaviour as on Wikipedia. This suggests that the diversity in possible mails, \textit{i.e.} the spread of distribution, is large enough to enable SPOT to discern human made mails from LLM ones. However, we remark that the results are less stable across different LLM architectures, as Falcon 7B and GPT-J 6B have a harder time classifying text inputs. This phenomenon cannot be attributed to their higher performance in terms of perplexity, as they are both outperformed by Mistral 7B. Similarly, we can rule out the explanation regarding training data, as none of these models were specifically trained on such messages. The reason as to why the difference in performance between models remains an open question to us. Still, SPOT manages to attribute an originality scores $8.53\times$ to $119.22\times$ larger to humans than to LLMs. Consequently, we suggest considering SPOT for practical applications. However, such applications often involve neural network compression, which we have not considered so far.
\begin{table}[!t]
\caption{Evaluation on actual emails sent and received by workers.}
\label{tab:mail}
    \centering
    \setlength\tabcolsep{4pt}
    \begin{tabular}{c|c|c|c|c|c|c|c}
    \hline
        ~ & human & Llama 7B & Falc 7B & Dv2 3B & GPT-J 6B & Mistral 7B & OPT 6.7B \\
        \hline
          llama 7B  &     1.073     &  0.004   &   0.002   &   0.005    &  0.005   &   0.007    &  0.009 \\
 falcon 7B  &     0.307     &  0.007   &   0.000   &   0.005    &  0.002   &   0.006    &  0.036 \\
 dollyv2 3B &     0.527     &  0.034   &   0.005   &   0.003    &  0.005   &   0.027    &  0.024 \\
  GPT-J 6B  &     0.349     &  0.019   &   0.003   &   0.005    &  0.002   &   0.016    &  0.003  \\
 Mistral 7B &     1.057     &  0.007   &   0.002   &   0.005    &  0.003   &   0.007    &  0.011  \\
  OPT 6.7B  &     0.394     &  0.024   &   0.002   &   0.005    &  0.004   &   0.016    &  0.001 \\
    \hline
    \end{tabular}
\end{table}

\subsection{Robustness to Deployment Conditions}
In Table \ref{tab:compressed}, we evaluate Llama models compressed using the popular OPTQ \cite{frantar2022optq} quantization technique. This method depends on two hyperparameters: the bit-width (which defines the compression rate) and the group-size (the larger the group size, the higher the fidelity to the uncompressed model). On Wikipedia, we observe that the larger the group-size, the closer the results are to the original Llama 7B model. Furthermore, it appears that SPOT is robust to compression on either the detection LLM (rows) or the LLM generator (columns). This final result leads us to conclude that SPOT is deployment friendly and compatible with quantization.

\begin{table}[!t]
\caption{Evaluation of the proposed originality score of compressed models on Wikipedia. LLMs are based on Llama 7B and fine-tuned using \cite{adaptllm} and compressed using OPTQ with bit-width (B) and groupings (G) parameters.}
\label{tab:compressed}
    \centering
    \setlength\tabcolsep{4pt}
    \begin{tabular}{c|c|c|c|c|c|c|c|c}
    \hline
        ~ & human & B4 128G & B4 32G & B4 64G & B8 1G & B8 32G & B8 128G & llama 7B \\ 
        \hline
    \hline
    B4 32G & \textbf{1.023} & 0.007 & \textit{0.008} & 0.007 & 0.007 & 0.007 & 0.007 & 0.007\\
    B4 64G & \textbf{1.023} & 0.005 & \textit{0.006} & 0.005 & 0.005 & 0.005 & 0.005 & 0.006\\
    B4 128G & \textbf{1.543} & 0.007 & \textit{0.009} & 0.007 & 0.007 & 0.007 & 0.007 & 0.008\\
    B8 1G & \textbf{1.294} & 0.005 & 0.006 & 0.005 & 0.005 & 0.005 & 0.005 & \textit{0.006}\\
    B8 32G & \textbf{1.410} & 0.005 & \textit{0.006} & 0.005 & 0.005 & 0.005 & 0.005 & 0.006\\
    B8 128G & \textbf{1.603} & 0.005 & \textit{0.006} & 0.005 & 0.005 & 0.005 & 0.005 & 0.006\\
    llama 7B & \textbf{1.754} & 0.007 & \textit{0.008} & 0.008 & 0.008 & 0.008 & 0.008 & 0.007\\
    \hline
    \end{tabular}
\end{table}

\section{Limitations}
As it stands, our study suffers from a few limitations. First, in order to evaluate multiple models across several tasks and applications, we focused on models with up to 7 billion parameters. While, we provided some results to suggest a robustness to architecture size, the study would benefit from an evaluation against significantly larger LLMs such as Gemini ultra. In particular, large models trained on unknown data, such as GPT 3.5 and GPT 4. A second shortcoming, which was thoroughly studied in \cite{chen2023can}, is the use case of human text slightly altered by an LLM. In our experimentation, the evaluated tokens were systematically 100\% human made or 100\% LLM generated. However, in practice, targeted attacks may include a mixture of human made text slightly edited by an LLM. We have not evaluated SPOT in such setups. Furthermore, regarding realistic setups, we know that compression can significantly alter models performance and behaviour. Currently, we only focused on quantization and did not consider sparsity and many other compression techniques to ensure good performance under deployment conditions.

\section{Conclusion}
In this study, we introduced a novel method, dubbed SPOT, to classify whether the source of text is an LLM or not. Our classifier relies on two emergent properties from LLMs. First, LLMs tend to learn similar patterns and probability sets across, enabling us to recognize different LLMs from a single reference one. This is our first new finding which we empirically validate on a wide range of subjects (medicine, coding, emails, Wikipedia, ...) and models from OPT to LLama at different model sizes the results remain robust. Second, LLM-based text can be distinguished from natural text simply based on the ability to reconstruct said text using an LLM. This property is at the core of the proposed SPOT method. Although the proposed method has limitations, we believe that it paves the way towards new safety protocols against malevolent LLM-generated content.

{\small
\bibliographystyle{unsrt}
\bibliography{neurips_2023.bib}
}
\newpage
\appendix
\section{STD}\label{sec:appendix_std}
\begin{figure}[!t]
    \centering
    \includegraphics[width = 0.6 \linewidth]{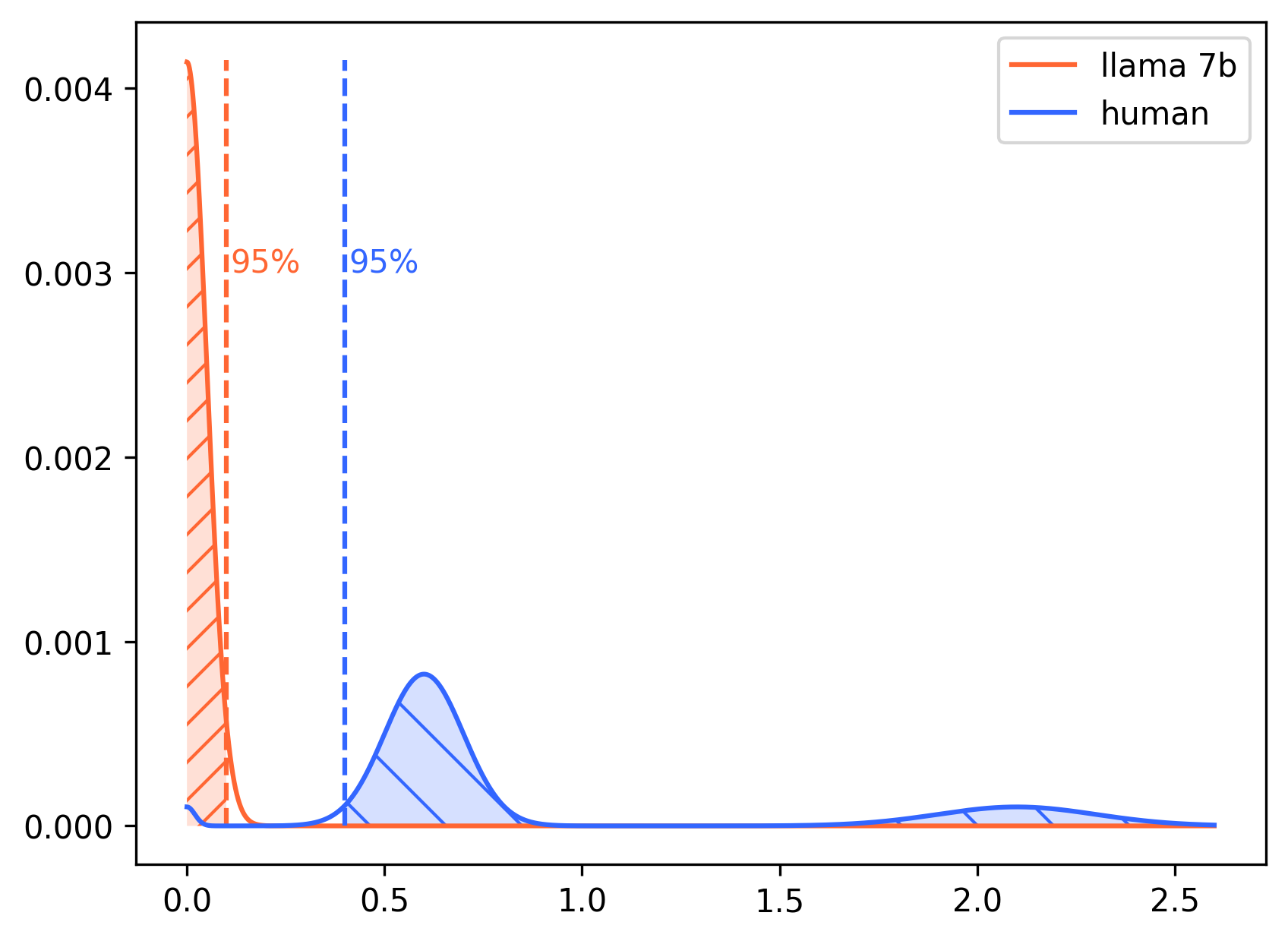}
    \caption{Distribution density of the originality score $\mathcal{O}$ for a subset of 10000 training examples from Wikipedia for both human and Llama 7B evaluated by Falcon 7B, with a context size of 24 tokens and 128 tokens long samplings. The distributions are obtained with Gaussian kernels of std $0.001$. We report the threshold for 95\% of the scores' threshold, which highlights the clear split between human and LLMs writing through the proposed metric.}
    \label{fig:confidence}
\end{figure}

\begin{figure}[!t]
    \centering
    \includegraphics[width = 0.6 \linewidth]{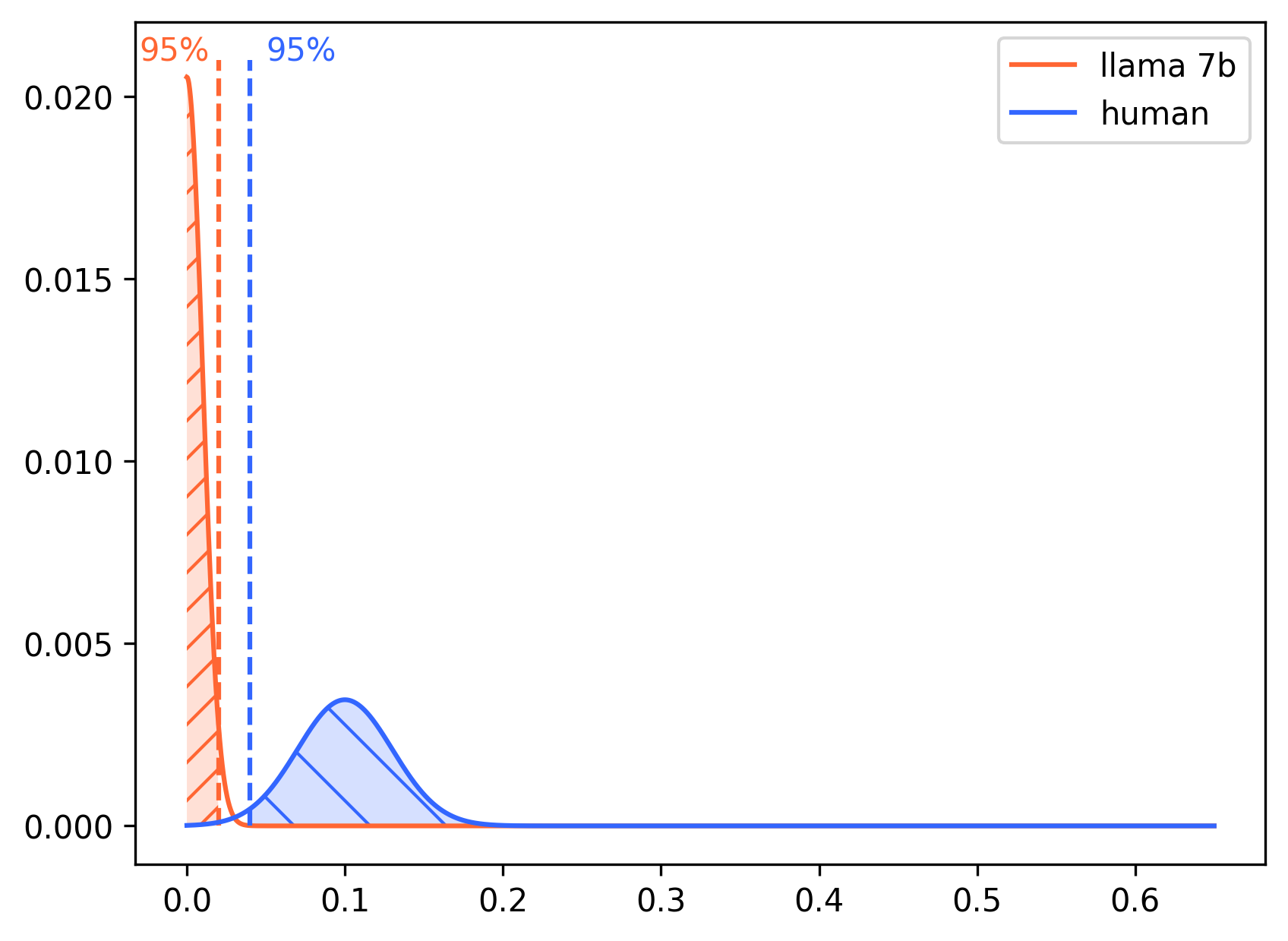}
    \caption{Distribution density of the originality score $\mathcal{O}$ for a subset of 10000 training examples from Wikipedia for both human and Llama 7B evaluated by Falcon 7B, with a context size of 24 tokens and 768 tokens long samplings.}
    \label{fig:confidence_further_gen}
\end{figure}

\begin{figure}[!t]
    \centering
    \includegraphics[width = 0.6 \linewidth]{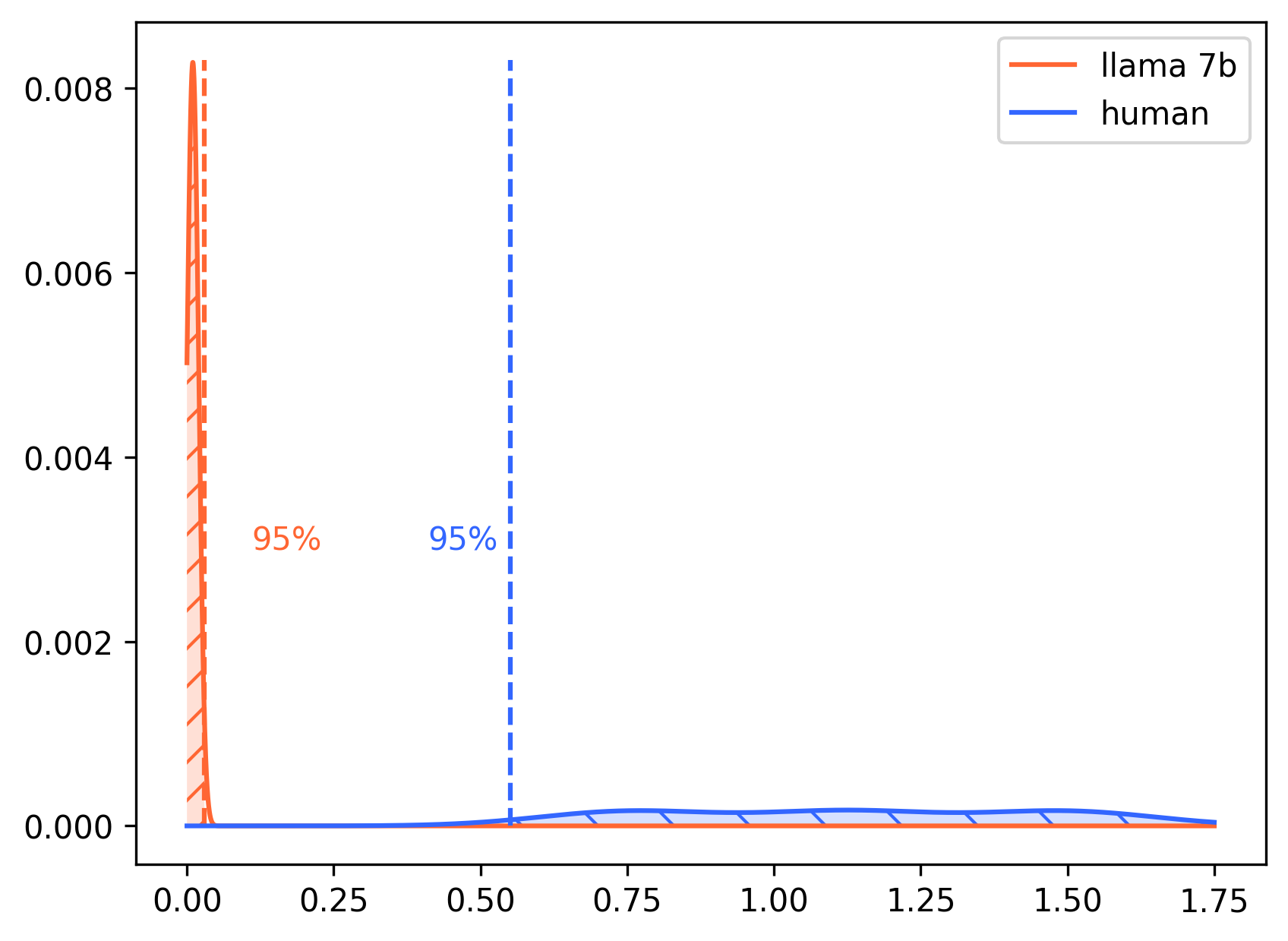}
    \caption{Distribution density of the originality score $\mathcal{O}$ for a subset of 10000 training examples from Wikipedia for both human and Llama 7B evaluated by Falcon 7B, with a context size of 512 tokens and 768 tokens long samplings.}
    \label{fig:confidence_long_context}
\end{figure}

In Figure \ref{fig:confidence}, \ref{fig:confidence_further_gen} and \ref{fig:confidence_long_context}, we plot the smoothed (using Gaussian kernels) distribution densities of the originality scores obtained on Wikipedia using a Llama 7B evaluated by a Falcon 7B. We observe the significant gap between the human and LLM samples. These plots highlight the relevance of equation \ref{eq:decision} by showing a clear frontier between the two classes through the lens of SPOT.

In particular, in Figure \ref{fig:confidence_further_gen}, we observe that on longer generations and text samples, the task of separating human from LLMs grows more challenging. This phenomenon can be explained by the divergence between LLMs on longer sequences from a shorter context, which increases the score they attribute to one another as compared to the human texts. Still, the separation enables us to achieve well over 95\% accuracy when trying to classify LLMs from human generated text samples.

On longer contexts (Figure \ref{fig:confidence_long_context}), we observe a massive increase in the variance of the originality score attributed to human. Still, the gap to LLM scores is not affected, thus leading to a simple task of origin classification.

\section{Extremely Short Generation}\label{sec:appendix_extremly_short}
In Table \ref{tab:wiki_extremly_short}, we evaluate the proposed SPOT method on 2 tokens generated from a context size of 24 tokens. The purpose of this experiment is to evaluate the robustness to very short completion in a challenging domain (Wikipedia). Contrary to specific domains such as mathematics (see Appendix \ref{sec:appendix_maths}), it appears that SPOT can detect very short LLM completions.
\begin{table}[!t]
\caption{Evaluation on extremely short generations of 2 tokens}
\label{tab:wiki_extremly_short}
    \centering
    \setlength\tabcolsep{1pt}
    \begin{tabular}{c|c|c|c|c|c|c|c|c|c|c}
    \hline
        ~ & human & O-LLa 3B & LLa 7B & Falc 7B & Dv2 3B & OPT 350M & OPT 2.7B & OPT 6.7B & GPT-J 6B & Mist 7B \\
        \hline\hline
O-LLa 3B & \textbf{1.583} & 0.157 & 0.158 & 0.201 & 0.006 & 0.005 & 0.005 & 0.004 & 0.028 & \textit{0.213}\\
LLa 7B & \textbf{1.781} & \textit{0.176} & 0.107 & 0.022 & 0.013 & 0.006 & 0.005 & 0.004 & 0.048 & 0.173 \\
Falc 7B & \textbf{0.762} & \textit{0.146} & 0.120 & 0.000 & 0.006 & 0.011 & 0.005 & 0.005 & 0.037 & 0.127\\
Dv2 3B & \textbf{1.043} & \textit{0.247} & 0.221 & 0.211 & 0.000 & 0.019 & 0.012 & 0.009 & 0.017 & 0.197\\
OPT 350M & \textbf{0.848} & 0.288 & 0.342 & 0.089 & 0.054 & 0.004 & 0.003 & 0.006 & 0.060 & \textit{0.305}\\
OPT 2.7B & \textbf{0.846} & 0.188 & 0.217 & 0.076 & 0.020 & 0.010 & 0.002 & 0.002 & 0.028 & \textit{0.193}\\
OPT 6.7B & \textbf{0.846} & 0.160 & \textit{0.174} & 0.073 & 0.018 & 0.007 & 0.002 & 0.001 & 0.020 & 0.158\\
GPT-J 6B & \textbf{0.919} & \textit{0.207} & 0.179 & 0.108 & 0.012 & 0.006 & 0.004 & 0.003 & 0.001 & 0.193\\
Mist 7B & \textbf{1.723} & \textit{0.162} & 0.151 & 0.079 & 0.009 & 0.008 & 0.003 & 0.001 & 0.099 & 0.140\\
    \hline
    \end{tabular}
\end{table}

\section{Evaluation on Mathematics}\label{sec:appendix_maths}
Extra results on mathematical prompting can be found in Table \ref{tab:math_algebra}, \ref{tab:math_calculus}, \ref{tab:math_proba} and \ref{tab:math_equalities}.
\begin{table}[!t]
\caption{Evaluation on linear algebra.}
\label{tab:math_algebra}
    \centering
    \setlength\tabcolsep{4pt}
    \begin{tabular}{c|c|c|c|c|c|c|c}
    \hline
        ~ & human & Llama 7B & Falc 7B & Dv2 3B & GPT-J 6B & Mistral 7B & OPT 6.7B \\ 
        \hline
    \hline
        llama 7B & 0.004 & 0.000 & 0.000 & \textit{0.019} & 0.001 & \textbf{0.046} & 0.003 \\
        falcon 7B & 0.015 & 0.001 & 0.000 & \textit{0.019} & 0.004 & \textbf{0.043} & 0.001 \\
        dollyv2 3B & \textit{0.018} & 0.007 & 0.013 & 0.001 & 0.002 & \textbf{0.200} & 0.001 \\
        GPT-J 6B & \textit{0.027} & 0.002 & 0.005 & 0.006 & 0.000 & \textbf{0.030} & 0.000 \\
        Mistral 7B & 0.002 & 0.001 & 0.000 & \textit{0.014} & 0.005 & \textbf{0.068} & 0.001 \\
        OPT 6.7B & \textbf{0.035} & 0.023 & 0.011 & 0.020 & 0.007 & \textit{0.025} & 0.000 \\
    \hline
    \end{tabular}
\end{table}
\begin{table}[!t]
\caption{Evaluation on calculus}
\label{tab:math_calculus}
    \centering
    \setlength\tabcolsep{4pt}
    \begin{tabular}{c|c|c|c|c|c|c|c}
    \hline
        ~ & human & Llama 7B & Falc 7B & Dv2 3B & GPT-J 6B & Mistral 7B & OPT 6.7B \\ 
        \hline
    \hline
        llama 7B & \textit{0.002} & 0.000 & 0.000 & 0.002 & 0.001 & \textbf{0.051} & 0.002\\
        falcon 7B & \textit{0.020} & 0.001 & 0.000 & 0.004 & 0.005 & \textbf{0.060} & 0.003\\
        dollyv2 3B & \textit{0.024} & 0.011 & 0.022 & 0.000 & 0.001 & \textbf{0.256} & 0.001\\
        GPT-J 6B & \textit{0.028} & 0.002 & 0.005 & 0.001 & 0.000 & \textbf{0.032} & 0.000\\
        Mistral 7B & 0.001 & 0.002 & 0.000 & 0.002 & \textit{0.003} & \textbf{0.052} & 0.001\\
        OPT 6.7B & 0.028 & \textbf{0.032} & 0.009 & 0.003 & 0.002 & \textit{0.029} & 0.000\\
    \hline
    \end{tabular}
\end{table}
\begin{table}[!t]
\caption{Evaluation on equlity assertions}
\label{tab:math_equalities}
    \centering
    \setlength\tabcolsep{4pt}
    \begin{tabular}{c|c|c|c|c|c|c|c}
    \hline
        ~ & human & Llama 7B & Falc 7B & Dv2 3B & GPT-J 6B & Mistral 7B & OPT 6.7B \\ 
        \hline
    \hline
        llama 7B & \textit{0.006} & 0.000 & 0.000 & 0.002 & 0.001 & \textbf{0.027} & 0.002 \\
        falcon 7B & 0.004 & 0.000 & 0.000 & 0.001 & \textit{0.009} & \textbf{0.010} & 0.002 \\
        dollyv2 3B & 0.000 & 0.001 & 0.001 & 0.000 & \textit{0.005} & \textbf{0.126} & 0.001 \\
        GPT-J 6B & \textit{0.002} & 0.000 & 0.001 & 0.002 & 0.000 & \textbf{0.006} & 0.001 \\
        Mistral 7B & \textbf{0.014} & 0.002 & 0.000 & 0.001 & \textit{0.009} & 0.004 & 0.001 \\
        OPT 6.7B & \textit{0.009} & 0.009 & 0.002 & 0.002 & 0.008 & \textbf{0.017} & 0.000 \\
    \hline
    \end{tabular}
\end{table}
\begin{table}[!t]
\caption{Evaluation on probabilities}
\label{tab:math_proba}
    \centering
    \setlength\tabcolsep{4pt}
    \begin{tabular}{c|c|c|c|c|c|c|c}
    \hline
        ~ & human & Llama 7B & Falc 7B & Dv2 3B & GPT-J 6B & Mistral 7B & OPT 6.7B \\ 
        \hline
    \hline
        llama 7B & 0.002 & 0.004 & 0.000 & 0.005 & \textit{0.005} & \textbf{0.023} & 0.005 \\
        falcon 7B & \textit{0.010} & 0.001 & 0.000 & 0.004 & 0.002 & \textbf{0.008} & 0.003 \\
        dollyv2 3B & \textbf{0.201} & 0.039 & 0.034 & 0.000 & 0.001 & \textit{0.048} & 0.005 \\
        GPT-J 6B & \textbf{0.038} & 0.003 & 0.004 & 0.005 & 0.000 & \textit{0.006} & 0.001 \\
        Mistral 7B & \textbf{0.025} & 0.003 & 0.000 & \textit{0.003} & 0.001 & 0.001 & 0.002 \\
        OPT 6.7B & \textbf{0.035} & \textit{0.027} & 0.022 & 0.010 & 0.000 & 0.008 & 0.000 \\
    \hline
    \end{tabular}
\end{table}

\section{Extreme Examples}
In this section, we simply list some edge case examples for SPOT on Llama 7B evaluated with Falcon. The following are examples generated with Llama, that are the likeliest to appear human to SPOT:
\begin{itemize}
\item"Martyn Rady (born 1955) is Professor of Central European History at the School of Slavonic and East European Studies, University College London. He is the author of The Czechoslovak Republic, 1918-1925: The Politics of the Weak"
\item "Switched is an American TV series that ran from 2003 through 2004 in whichthe main character, a young man named Eddie, is a time traveler who uses his knowledge of the future to make money.$\backslash$ nThe series was created by David S. Goyer and Bran"
\item "The Kosciusko School District is a public school district based in Kosciusko, Mississippi.$\backslash$ n$\backslash$ nSchoolsin the district include:$\backslash$ n$\backslash$ n* Kosciusko Elementary School$\backslash$ n* Kosciusko Middle School$\backslash$ n* Kosciusko High School$\backslash$ n$\backslash$ n\#\# External links$\backslash$ n$\backslash$ n* Kosciusko School"
\item "Brunner's glands (or duodenal glands) are compound tubular submucosalglands that are located in the duodenum. They are found in the duodenum of all mammals, including humans. They are also found in the duodenum of birds,"
\item "The Layene (also spelled Layène, Layenne, or Layeen) is atown and commune in the Khenchela Province of Algeria. ... The Layene (also spelled Layène, Layenne, or Layeen) is a town"
\item "Aramac Airport  is an unlicensed airport located  from the town of Aramac in remotenorthwestern Michigan.$\backslash$ n$\backslash$ n\#\# External links$\backslash$ n$\backslash$ n* Aerial image as of 1996 from USGS The National Map$\backslash$ n* Resources for this airport:$\backslash$ n  * F"
\item "Adam’s Task: Calling Animals by Name by philosopher, poet, and animal trainer Vicki Hearne.$\backslash$ nThe book is a collection of essays, some of which are about Hearne’s experiences with animals, and others about the philosophy of animals. The essays"
\item "The Multi-Application Survivable Tether (MAST) experiment was an in-space investigation designed to usethe Space Shuttle to demonstrate the ability to deploy and retrieve a tether in space. The experiment was conducted in the Shuttle's payload bay and used the Shuttle's"
\item "WOW Hits 2001 is a compilation album of thirty Contemporary Christian music hits and three bonustracks. The album was released on October 23, 2001 by WOW Hits.$\backslash$ n$\backslash$ nProfessional ratings$\backslash$ n$\backslash$ n\#\# Tracklisting$\backslash$ n$\backslash$ n1."
\item "Mycobacterium malmoense is a Gram-positive bacterium from the genus Mycobacterium which was isolated from the blood of a patient with a mycobacterial infection. \# 2019–2020-as UEFA-bajn"
\end{itemize}

To the contrary, we list examples that were written by people but may look like they were LLM generated:
\begin{itemize}
    \item "Moss Mabry (July 5, 1918 – January 25, 2006) was an American costume designer.$\backslash$n$\backslash$nBiography$\backslash$n$\backslash$nHe started off designing costumes for his high school plays, but actually studied mechanical engineering at the University of Florida.",
    \item "Xu Caihou (; June 1943 – March 15, 2015) was a Chinese general in the People\'s Liberation Army (PLA) and vice-chairman of the Central Military Commission (CMC), the country\'s top military council. As Vice-chairman of the CMC, he was one of the top ranking officers of the People\'s Liberation Army.",
    \item "Parvati (, ), Uma (, ) or Gauri (, ) is the Hindu Goddess of power, nourishment, harmony, devotion, and motherhood. She is Devi in her complete form. Along with Lakshmi and Saraswati, she forms the Tridevi.$\backslash$n$\backslash$nParvati is the wife of the Hindu god Shiva.",
    \item "Tha Bo (, ) is a district (amphoe) in the western part of Nong Khai province, northeastern Thailand.$\backslash$n$\backslash$nGeography$\backslash$nNeighboring districts are (from the east clockwise): Mueang Nong Khai and Sa Khrai of Nong Khai Province; Ban Phue of Udon Thani province; Pho Tak and Si Chiang Mai of Nong Khai Province.",
    \item "02, O-2, o2, O2, O2 or O2- may refer to:$\backslash$n$\backslash$nScience and technology $\backslash$n$\backslash$n  or dioxygen, the common allotrope of the chemical element oxygen$\backslash$n , the ion oxide$\backslash$n , the ion superoxide$\backslash$n , the ion dioxygenyl$\backslash$n , doubly ionized oxygen$\backslash$n O2, an EEG electrode site according to the 10–20 system",
    \item "Felsőregmec () is a village in the Borsod-Abaúj-Zemplén county in northeastern Hungary.$\backslash$n$\backslash$nExternal links $\backslash$n Street map $\backslash$n$\backslash$nPopulated places in Borsod-Abaúj-Zemplén County$\backslash$nRomanesque architecture in Hungary",
    \item "Sisowath Monivong (,  ; 27 December 1875 – 24 April 1941) was the King of Cambodia from 9 August 1927 until his death in 1941. During his reign, Cambodia was a French protectorate.",
    \item "Kuppali, also known as Kuppali, is a small village in Thirthahalli taluk of Shimoga district in the state of Karnataka in India. It is famous for being the childhood home of the renowned Kannada poet Kuvempu.",
    \item "The banns of marriage, commonly known simply as the 'banns' or 'bans'  (from a Middle English word meaning 'proclamation', rooted in Frankish and from there to Old French), are the public announcement in a Christian parish church or in the town council",
    \item "2010 was a year.$\backslash$n$\backslash$n2010 may also refer to:$\backslash$n2010s, the decade$\backslash$n2010 FIFA World Cup, the 19th FIFA World Cup, the world championship for men's national association football teams$\backslash$n2010 Winter Olympics, February",
    
\end{itemize}

\end{document}